\documentclass[twoside,leqno,twocolumn]{article}

\usepackage[letterpaper]{geometry}

\usepackage{ltexpprt}
\usepackage{hyperref}

\usepackage{romannum}
\usepackage{amsmath}
\usepackage{multirow}
\usepackage{algorithm}
\usepackage{algorithmic}
\usepackage{caption}
\usepackage{subcaption}
\usepackage{colortbl}
\usepackage{amssymb}
\usepackage{graphicx}
\usepackage{float}
\usepackage{wrapfig}

\usepackage{tikz}
\newcommand*{\circled}[1]{\lower.7ex\hbox{\tikz\draw (0pt, 0pt)%
    circle (.5em) node {\makebox[1em][c]{\small #1}};}}
    
\usepackage{url}            
\usepackage{booktabs}       
\usepackage{amsfonts}       
\usepackage{nicefrac}       
\usepackage{microtype}      

\usepackage{xcolor}         
\DeclareMathOperator{\sign}{sign}

\begin{document}

\newcommand\relatedversion{}
\renewcommand\relatedversion{\thanks{The full version of the paper can be accessed at \protect\url{https://arxiv.org/abs/1902.09310}}} 


\title{EsaCL: An Efficient Continual Learning Algorithm}
\date{}

\author{Weijieying Ren \\
  The Pennsylvania State University \\
  State College, Pennsylvania\\
  \texttt{wjr5337@psu.edu}
  \and
  Vasant G Honavar 
  \footnotemark[1]\\
  The Pennsylvania State University \\
  State College, Pennsylvania\\
  \texttt{vuh14@psu.edu}
}

\maketitle
\footnotetext[1]{Corresponding author}
\fancyfoot[R]{\scriptsize{Copyright \textcopyright\ 2024 by SIAM\\
Unauthorized reproduction of this article is prohibited}}


\fancyfoot[R]{\scriptsize{Copyright \textcopyright\ 2024\\
Copyright retained by principal author's organization.}}





\begin{abstract}

A key challenge in the continual learning setting is to efficiently learn a sequence of tasks without forgetting how to perform previously learned tasks. Many existing approaches to this problem work by either retraining the model on previous tasks or by expanding the model to accommodate new tasks. However, these approaches typically suffer from increased storage and computational requirements, a problem that is worsened in the case of sparse models due to need for expensive re-training after sparsification. To address this challenge, we propose a new method for efficient continual learning of sparse models (EsaCL) that can automatically prune redundant parameters without adversely impacting the model's predictive power, and circumvent the need of retraining. We conduct a theoretical analysis of loss landscapes with parameter pruning, and design a directional pruning (SDP) strategy that is informed by the sharpness of the loss function with respect to the model parameters. SDP ensures model with minimal loss of predictive accuracy, accelerating the learning of sparse models at each stage. To accelerate model update,  we introduce an intelligent data selection (IDS) strategy that can identify critical instances for estimating loss landscape, yielding substantially improved data efficiency. The results of our experiments show that EsaCL achieves performance that is competitive with  the state-of-the-art methods on three continual learning benchmarks, while using substantially reduced memory and computational resources. 

\end{abstract}
\section{Introduction}
Many real-world applications require the system to learn a sequence of tasks \cite{de2021continual,gurbuz2022nispa} over time in such a manner that learning of new tasks does not result in catastrophic forgetting leading to significant degradation in performance on the previously learned tasks. Much recent work on continual learning has focused on strategies for avoiding catastrophic forgetting, e.g., gradually expanding the capacity of the model \cite{golkar2019continual,gurbuz2022nispa} to accommodate new tasks, or retraining the model on previously learned tasks while being trained on new tasks through ``replay" methods  \cite{chaudhry2019tiny, buzzega2020dark,rebuffi2017icarl,rolnick2019experience}. However, these approaches typically incur increased memory and computational
overhead \cite{wangsparcl}. Hence, there is an urgent need for methods that reduce the memory, computation, and training data needs of continual learning, while avoiding significant drop in performance. 

Against this background, we propose EsaCL which eliminates  the extensive retraining phase and avoids the need to search for an optimal sparsity threshold. It further improves learning efficiency through an intelligent selection of informative training data samples. Specifically, we introduce two novel strategies to improve the efficiency of training in the continual learning setting: 
(1) EsaCL searches for a region of the objective function that is relatively flat in the neighborhood of its minimum. We set the optimization direction by regularizing back-propagated gradients, so as to guide model to converge to a point lying in the flat region of the objective function \cite{foretsharpness} while driving the less important weights towards zero. In the flat portion of the objective function, the model would be relatively insensitive to small perturbations in the weights. The sparsity automatically achieved in learning with a K-sparse polytope based optimization strategy. (2) EsaCL selects a  subset of training examples that suffice for achieving the desired performance so as to reduce the computational cost of learning.  Our main contributions are as follows:

\vspace*{5pt}
\begin{itemize}
\vspace*{-0.1in}
\item
To the best of our knowledge, this work represents the first of its kind for learning a sparse model with one-shot pruning without re-training with resulting improvements in terms of model size as well as computational cost. Specifically, we present the first approach that exploits flat region of the objective function to obtain an efficient algorithm for continual learning.
The key elements of the proposed solution include: a sharpness-sensitive pruning strategy and an efficient data selection method, both of which contribute to improvements in efficiency of EsaCL without degradation in performance relative to SparCL.
\vspace*{-0.1in}
\item
We present results of extensive experiments on standard continual learning benchmark data sets that show that EsaCL consistently achieves performance that is competitive with the the state-of-the-art approaches over a wide range of pruning ratios without retraining,  while using substantially reduced memory and computational resources.
\end{itemize}
\vspace*{-0.1in}
The rest of the paper is organized as follows: Section 2 reviews related work. Section 3 introduces our problem formulation and  describes the EsaCL, the proposed efficient continual learning algorithm. Section 4 describes the experimental setup and results. Section 5 concludes with a summary and brief discussion.
\section{Related Work}
\subsection{Continual Learning} 
Continual learning focuses on sequential learning of new tasks, ideally without access to historical data, while leveraging previously gained knowledge \cite{zhao2020balancing}. A significant challenge in continual learning has to do with avoiding catastrophic forgetting, which inevitably leads to a drop in performance on previously learned tasks when trained on new tasks.
Existing approaches to continual learning can be broadly classified into three categories: 1) Rehearsal-based methods which reload historical data \cite{chaudhry2019tiny} or use synthetic data generated from a generative model trained on historical data to circumvent catastrophic forgetting.
2) Regularization-based methods \cite{li2017learning} penalize model parameter change to mitigate catastrophic forgetting as new tasks are learned; 3) Parameter isolation methods mitigate catastrophic forgetting by allocating different subsets of parameters to different tasks, thereby minimizing the interaction between tasks \cite{kang2022forget,zhao2022exploring}. 

Despite the recent advances in CL (surveyed in \cite{de2021continual,wang2023comprehensive}), there has been limited attention paid to computational, data, memory, and model efficiency considerations that are critical for large-scale applications of CL. 
SparCL \cite{wangsparcl} pioneered an approach to address model efficiency in the CL setting by inducing model sparsity using $l_1$ regularization. While SparCL succeeds in avoiding catastrophic forgetting, it does so at considerably increased computational overhead which grows with the number of tasks. Furthermore, it's applicability in real-world settings is limited by its need for extensive trial and error for setting the pruning thresholds. SupSup \cite{golkar2019continual} utilizes task-specific masks to isolate parts of the network that are modified during learning to mitigate catastrophic forgetting but at the expense of substantial increase in model size \cite{kang2022forget}. EsaCL on the other hand aims to perform single-shot model compression, without the need for expensive retraining after each compression step, while preserving the model's generalization ability.
\subsection{Sharpness-aware Minimization}
 Hochreiter and Schmidhuber \cite{hochreiter1994simplifying} were the first to introduce a method to search for a 'flat' minimum of a neural network objective function to improve generalization. Subsequent work, such as that of Foret et al. \cite{foretsharpness}, explored the relationship between loss values and the loss landscape, and found evidence to suggest that flat minima enhance the generalization ability of the models.  Drawing  inspiration from the work of Keskar et al. \cite{keskar2016large}, several sharpness-Aware Minimization (SAM) algorithms have been introduced, including those by Foret et al. \cite{foretsharpness}. Related work has focused on reducing the computational cost of SAM \cite{liu2022towards,foretsharpness,zhao2020semi} using  gradient decomposition \cite{liu2022towards}, and objective function approximation. Recent work has used SAM for continual learning \cite{kim2022fisher}, cross-domain knowledge transfer, and pseudo-label generation in semi-supervised learning.
In light of this body of work, the key contribution of this paper is to connect SAM to a pruning based approach to continual learning. Specifically, we use SAM to devise an efficient pruning method that increases model sparsity without degrading model performance; and use sample selection to improve the overall efficiency of the method.  


\section{Efficient Sharpness Aware Continual Learning}
\subsection{Continual learning} 
A CL problem can be entails learning from a sequentially ordered set of tasks $\{1,2,...,T\}$ where each task is specified by a multi-set of labeled samples, or input-label pairs. Thus, the $t$-th task is specified by $\mathcal{D}_t = \{(\textbf{x}_i,y_i)\}_{i=1}^{N_t}$, where $N_t$ represents the number of training examples for the  $t$-th task.
When learning task $t$, the algorithm is allowed access only to  data  $\mathcal{D}_t$. There are two variants of continual learning: In task incremental learning, the identities of the tasks are made available during testing. In  class incremental learning, the task identities are not available during testing.
Formally, the objective of CL is to learn function $f:\mathbb{R}^d\rightarrow \mathbb{R}$
with parameters $\boldsymbol{\theta} \in \mathbb{R}^d$ that minimizes the loss over the  tasks:
\begin{equation}
\operatorname*{min}_{\boldsymbol{\theta}} \mathbb{E}_{t=1}^{|T|}[\mathbb{E}_{(x,y)\sim \mathcal{D}_{t}}[\ell(f_{\boldsymbol{\theta}}(x),y)]],
\end{equation}
where $\ell$ is an appropriate loss, e.g., cross-entropy loss.
To accommodate model size constraints, we introduce $\mathcal{C}$ as a pruning operator for model parameters.
To improve data efficiency, we perform intelligent sample selection to choose a subset  $\mathcal{S}_t \sim \mathcal{D}_t$ for each task. This yields the following objective function:
\begin{equation}
\operatorname*{min}_{\boldsymbol{\theta}} \mathbb{E}_{t=1}^{|T|}[\mathbb{E}_{(x,y)
\sim 
\mathcal{S}_{t}}[\ell (f_{\mathcal{C}(\boldsymbol{\theta})};S_t)]].
\label{eq::overall_2}
\end{equation}
The goal of optimization is  to effectively learn new tasks without degradation in performance on previously learned tasks. Efficiency considerations call for achieving continual learning while minimizing the computational resources and memory used without incurring a significant drop in performance. We proceed to describe our approach to realizing these objectives.
\vspace*{-0.2cm}
\subsection{Rethinking iterative pruning in CL settings.}
As mentioned,  parameter magnitude based pruning methods in the CL setting rely on computationally expensive retraining of the model after each pruning step which increases in proportion to the number of tasks. Additionally, continual learners are sensitive to sparsity ratio, i.e., the fraction of parameters that are retained relative to the total number of parameters. Determining the optimal sparsity ratio requires either substantial domain knowledge or extensive trial-and-error \cite{hoefler2021sparsity,chao2020directional}. We proceed to address this challenge by introducing a novel one-shot pruning strategy that makes EsaCL computationally efficient and its performance relatively insensitive to the choice of the pruning ratio.
\begin{figure}
\begin{center}
\includegraphics[width=1.0\linewidth]{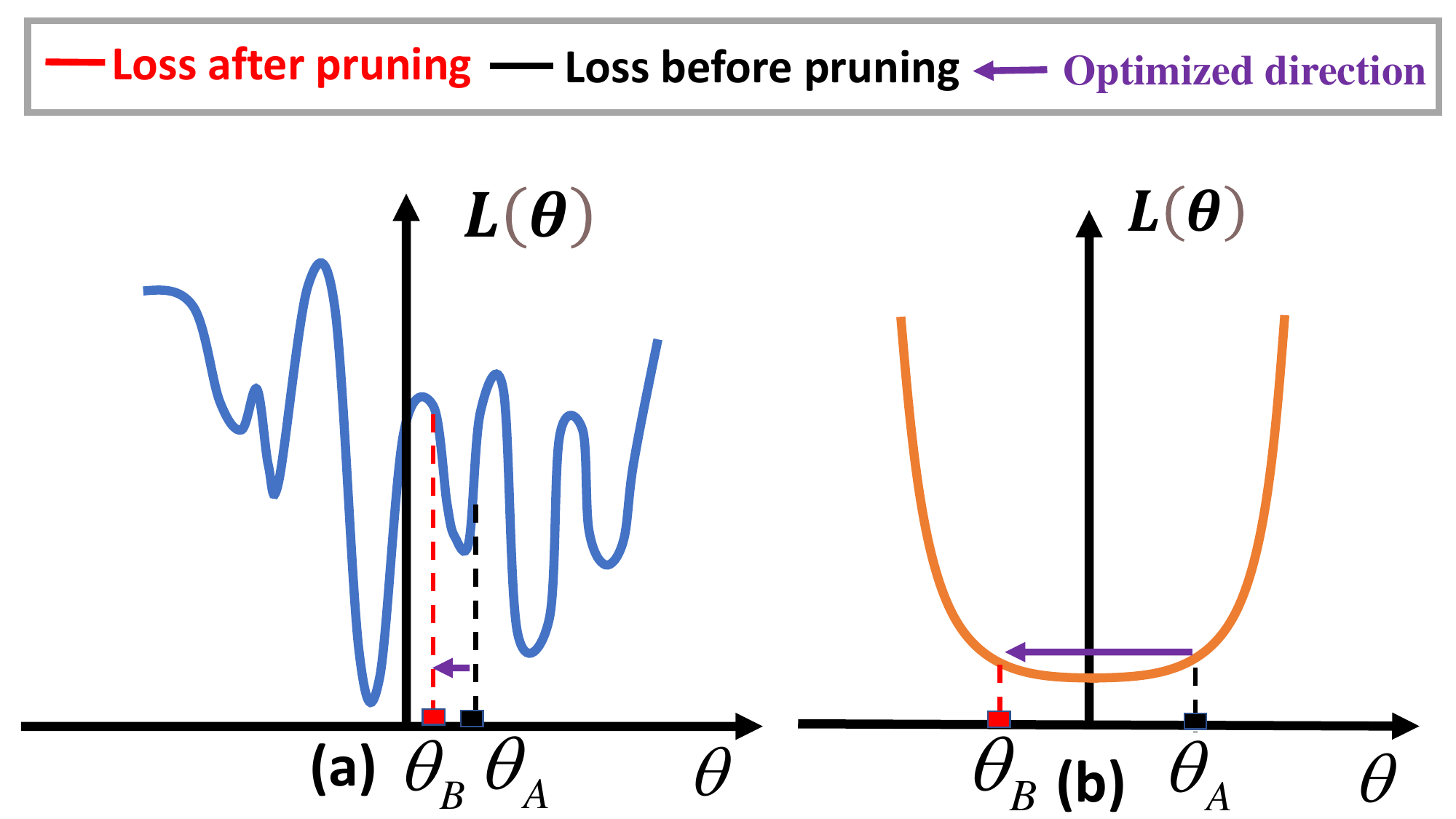}
  \end{center}
\caption{Illustration of why a flat loss landscape loss surface improve directional pruning without retraining.}
~\label{fig::motivation}
\vspace{-0.4cm}
\end{figure}
\vspace{-0.1cm}
We begin by investigating how the training loss changes as some of the model parameters $\boldsymbol{\theta}$ get pruned.
Such pruning can be implemented by applying a mask
$\boldsymbol{m} \in \{0,1\}$
to the weights, yielding a sparse model:
$\mathcal{C}(\boldsymbol{\theta}) = \boldsymbol{m} \odot \boldsymbol{\theta}$.
This can be viewed as a form of weight perturbation \cite{lin2020dynamic}:
\begin{equation}
\mathcal{C}(\boldsymbol{\theta}) = \boldsymbol{\theta} - A \cdot \sign(\boldsymbol{\theta}),
\end{equation}
where $A$ is a diagonal matrix with 
$0\leq A_{ii}\leq |\boldsymbol{\theta}_i| $ for $i = 1,2,...,d$. 
And 
$\sign(\boldsymbol{\theta}) \in \{-1,1\}^d$ 
is the sign vector of 
$\boldsymbol{\theta}$. 
Specifically, $i$-th coefficient is pruned if 
$A_{ii} = |\boldsymbol{\theta}_i|$. 
Model retraining is necessary since $\boldsymbol{\theta}-A \cdot \sign(\boldsymbol{\theta})$ does not lie in the null-space of weight parameters. The shape of the loss landscape is sensitive to parameter changes \cite{foretsharpness,liu2022towards}.  
As Figure. \ref{fig::motivation}(a) shows, the training loss changes drastically when the model parameters change from $\theta_A$ to $\theta_B$. Hence,  achieving satisfactory performance with parameters $\theta_B$ requires retraining the model.

Instead, we seek a flat region of the loss landscape $\mathcal{M}$ so as to  
minimize the difference in loss between $f(\boldsymbol{\theta};\mathcal{D}_t)$ and $f({\mathcal{C}(\boldsymbol{\theta})};\mathcal{D}_t)$. With a flattened loss landscape shown in Figure \ref{fig::motivation} (b),  the training loss remains relatively stable despite model pruning. In other words, $f({\mathcal{C}(\boldsymbol{\theta})};\mathcal{D}_t) \approx f(\boldsymbol{\theta};\mathcal{D}_t)$. Consequently, perturbing the model parameters minimally impacts the training loss with a flattened loss $\mathcal{M}$.
We can ensure this condition by finding $\boldsymbol{v}^*$ that approximates $\boldsymbol{\theta}$:
\begin{equation}
\boldsymbol{v}^* = \arg \min_{\boldsymbol{v} \in \mathcal{M}} ||\boldsymbol{v} - \sign(\boldsymbol{\theta})||_2^2,
\label{eq::optimal_v}
\end{equation}
i.e., $\boldsymbol{v}^*$ = $\mathcal{P}_{\mathcal{M}}(\sign(\boldsymbol{\theta}))$,
where $\mathcal{P}_{\mathcal{M}}$ is the projection operator on space $\mathcal{M}$.
As Equation \ref{eq::optimal_v}, shows, while
$\boldsymbol{v}^* \in \mathcal{M}$ 
approximates $\boldsymbol{\theta}$, it does not always decrease the magnitude of $\boldsymbol{\theta}$. It encourages sparsity of $\boldsymbol{\theta}$  when 
$\sign(\boldsymbol{v}_k)$ = $\sign(\boldsymbol{\theta}_k)$, that is:
\begin{equation}
s_k := \sign(\boldsymbol{\theta}_k) \cdot \mathcal{P}_{\mathcal{M}}(\sign(\boldsymbol{\theta})) > 0,
\label{eq::indicator_5}
\end{equation}
where $s_k$ induces sparsity by reducing the magnitude of the parameters along the $k$-th dimension with minimal impact on the the training loss. Equivalently, we constrain $\boldsymbol{\theta}$ to lie within a convex ball $\boldsymbol{\theta} \in \Pi(s_k)$, where coefficients marked by $s_k$ are expected to decay to 0. We achieve this by reformulating a CL-based pruning in Equation \ref{eq::overall_2} as follows:
\begin{equation}
\begin{aligned}
\operatorname*{min}_{\boldsymbol{\theta} \in \mathcal{M}} 
&\mathbb{E}_{t=1}^{|T|}[\mathbb{E}_{(x,y)
\sim 
\mathcal{S}_{t}}[\ell (f_{\mathcal{C}(\boldsymbol{\theta})};S_t)]]\\
\textrm{s.t.} 
&\quad \boldsymbol{\theta} \in {\Pi(s_j)}.
\label{eq::one_shot_pruning_6}
\end{aligned}
\end{equation}
Compared to Equation 2, the key difference here is to change the direction of the weight updates by taking into account both the loss landscape and the sparsity of the parameters. Additionally, we select informative  data samples $\mathcal{S}_t$ for each task to further reduce computational cost without degrading performance. The result is a sparse model that achieves the desired performance-efficiency trade-off.



We now proceed to turn the preceding intuition into an algorithmic solution for efficient continual learning. The key idea is to identify and exploit the flat region  $\mathcal{M}$ of the objective function given by Equation \ref{eq::one_shot_pruning_6}. We propose a two-part solution to achieving this goal: (1) sharpness-aware model pruning (section 3.3) and intelligent data selection (section 3.4). The first part sparsifies the model while incurring minimal changes in the training loss. The second selects informative data samples while avoiding significant drop in performance. 
The resulting EsaCL algorithm is shown in Algorithm 1.
\begin{algorithm}[ht]
\caption{Efficient Continual Learning Algorithm.}
    \label{alg:algorithm}
    \textbf{Input:} 
    Task sequences $\{\mathcal{D}_1,,,\mathcal{D}_T$\};
    Model weight $\boldsymbol{\theta}$;
    Batch size $b$;
    Learning rate $\eta$;
    Training iterations of $t$-th task $E_T$;
    Neighborhood size $\rho$ > 0;
    Sparsity ratio $s\%$.
    \begin{algorithmic}[1] 
        \FOR{task t = 1 , ... , $T$}
        \FOR{iteration $t$ = 1 , ... , $E_T$}
        \STATE Sample a mini-batch $\mathcal{B} \subset \mathcal{D}_t$ with batch size $b$; \\
        \STATE Compute loss values $\ell(\boldsymbol{\theta}_{t-1};\mathcal{B})$ and select support samples $\mathcal{B}_s$;
        \STATE Compute perturbation $\boldsymbol{\epsilon}_t = \rho \cdot\frac{\bigtriangledown_{\boldsymbol{\theta}} \ell(\boldsymbol{\theta}_{t-1};\mathcal{B}_s)}{||\bigtriangledown_{\boldsymbol{\theta}} \ell(\boldsymbol{\theta}_{t-1};\mathcal{B}_s)||_2}$
        \STATE Compute 
        $\ell(\boldsymbol{\theta}_{t-1}+\boldsymbol{\epsilon}_t;\mathcal{B}_s)$ in the outer minimization step
        and update gradient estimator $\bigtriangledown_{\boldsymbol{\theta}} \ell(\boldsymbol{\theta}_{t-1} + \boldsymbol{\epsilon}_t;\mathcal{B}_s)$
        \STATE Update momentum vectors 
                $\boldsymbol{m}_t\leftarrow \boldsymbol{m}_{t-1}+\alpha(\bigtriangledown_{\boldsymbol{\theta}} \ell(\boldsymbol{\theta}_{t-1} + \boldsymbol{\epsilon}_t;\mathcal{B}_s))$
        \STATE Solve linear minimization oracle (LMO): $\boldsymbol{v}_t\leftarrow LMO_c(\boldsymbol{m}_t)$
        \STATE Update neural network weight $\boldsymbol{\theta}_t\leftarrow \boldsymbol{\theta}_{t-1}+\eta\cdot(\boldsymbol{v}_t-\boldsymbol{\theta}_{t-1})$
        \ENDFOR

        \STATE Pruning neural network paramaters with sparsity ratio $s\%$
        
        \STATE Freeze the pruned neural network parameters and their connections: $\theta_\text{t}=\text{Freeze}(\theta_\text{t}$)
        
        \STATE Expand neural network parameters by reinitialize unused weights
        
        \ENDFOR   
        \RETURN

    \end{algorithmic}     
\end{algorithm}

\vspace{-0.8cm}
\subsection{Sharpness-aware informed Pruning.}
In order to progressively prune model parameters without re-training, we explicitly search for a feasible region $\mathcal{M}$ of the objective function given by Equation \ref{eq::one_shot_pruning_6} and project the gradient to a sparse subspace to sparsify the model without sacrificing its performance.
An intuitive approach to achieving this objective is to identify parameter values $\boldsymbol{\theta}$ that reside in the flat basin of the loss function where most of the eigenvalues in the Hessian matrix are close to zero. However, optimizing solely for the training loss could yield a model that fails to generalize beyond the training data.

Alternatively, we can utilize SAM (Sharpness-aware Minimization) \cite{foretsharpness,liu2022towards,kim2022fisher} to directly exploit the loss geometry to identify parameter values that are surrounded by neighborhoods where the training loss is small  and the loss landscape is flat. We can quantify the maximum change in empirical loss in response to a small perturbation $\boldsymbol{\epsilon}$ in the parameters $\boldsymbol{\theta}$ as follows:
\begin{equation}
\max_{\boldsymbol{\epsilon} :||\boldsymbol{\epsilon} ||_2< \rho }[\ell (\boldsymbol{\theta} + \boldsymbol{\epsilon};\mathcal{D}_t) 
-
\ell(\boldsymbol{\theta};\mathcal{D}_t)].
\label{eq::max_ep_7}
\end{equation}
Equation \ref{eq::max_ep_7} quantifies how much the training loss increases when $\boldsymbol{\theta}$ changes to a nearby parameter value $\boldsymbol{\theta}+\boldsymbol{\epsilon}$ in a Euclidean ball with radius
$\rho$. From equations \ref{eq::one_shot_pruning_6} and Equation \ref{eq::max_ep_7} 
we see that if the change in loss is less than or equal to a threshold value $\tau$:
$|| \ell (\boldsymbol{\theta} + \boldsymbol{\epsilon};\mathcal{D}_t)- \ell (\boldsymbol{\theta} ;\mathcal{D}_t)||_2 \leqslant \tau$, then the model parameters can be perturbed (i.e., be pruned) gradually. 
As Figure \ref{fig::motivation} (b) shows, the training loss changes little when we change the model parameter from $\theta_A$ to $\theta_B$.
We exploit the properties of SAM \cite{foretsharpness} to formulate the search for a flat region $\mathcal{M}$ of the objective function given by Equation \ref{eq::one_shot_pruning_6} as a min-max game \cite{foretsharpness} as follows:
\begin{equation}
\begin{aligned}
 \operatorname*{min}_{\boldsymbol{\theta}}\max_{\boldsymbol{\varepsilon }} 
 &\mathbb{E}_{t=0}^{|T|}\mathbb{E}_{\mathcal{D}_t}[\ell(f(\boldsymbol{\theta}+\boldsymbol{\varepsilon});\mathcal{D}_t)]\\
\textrm{s.t.} 
&\quad \boldsymbol{\theta} \in {\Pi(s_k)}.
\label{eq::one_shot_pruning_8}
\end{aligned}
\end{equation}

Equation \ref{eq::one_shot_pruning_8} aims to simultaneously minimize both the training loss and the sharpness of the loss. 
To sparsify $\boldsymbol{\theta}$, we assume that ${\Pi(s_k)}$ is a $K$-sparse polytope \cite{pokutta2020deep}.
Formally, the $K$-sparse polytope is obtained as the intersection of the $L1$-ball $\mathcal{B}1(\tau K)$ and the hypercube $\mathcal{B}{\infty}(\tau)$, where $\tau$ represents the polytope radius.
Without loss of generality, such a polytope can be defined as the convex hull spanned by all vectors in $\mathbb{R}^n$ with exactly $K$ non-zero entries, each of which is either $-\tau$ or $+\tau$.
In other words, we express the projection operator $\Pi(s_k)$ as a $K$-sparse polytope $\mathcal{C}(K,\tau)$.
Now we can rewrite Equation \ref{eq::one_shot_pruning_8} as follows:

\begin{equation}
\begin{aligned}
 \operatorname*{min}_{\boldsymbol{\theta}}\max_{\boldsymbol{\varepsilon }}
&\mathbb{E}_{t=0}^{|T|} \mathbb{E}_{\mathcal{D}_t}[\ell (f(\boldsymbol{\theta}+\boldsymbol{\varepsilon});\mathcal{D}_t)]\\
\textrm{s.t.} 
&\quad \boldsymbol{\theta} \in \mathcal{C}(K,\tau).
\label{eq::one_shot_pruning_9}
\end{aligned}
\end{equation}

\subsubsection{Optimization Process.} Optimization of Equation 9 involves two aspects: (1) Optimizing $\boldsymbol{\theta}$ to achieve a flat landscape by employing a min-max game, and (2) Projecting $\boldsymbol{theta}$ onto a sparse space. 
The first can be addressed using a simple approximation method, as in  SAM \cite{foretsharpness}. Hence we focus on the second aspect, namely, sparse optimization.
\vspace*{-0.1cm}
\subsubsection*{Projection-free Optimization.}
It is evident that Equation \ref{eq::one_shot_pruning_9} cane be viewed as projection gradient descent problem \cite{boyd2004convex}:
\begin{equation}
 {\Pi_{\mathcal{C}(K,\tau)}}(\operatorname*{min}_{\boldsymbol{\theta}}\mathbb{E}_{t=0}^{|T|}\max_{\boldsymbol{\varepsilon }} [\mathbb{E}_{\mathcal{D}_t}[\ell (f(\boldsymbol{\theta}+\boldsymbol{\varepsilon});\mathcal{D}_t)]]).
\end{equation}
However, there is a difficulty in that each projection step can be computationally expensive, and there is no closed expression for $\mathcal{C}(K,\tau)$.
Therefore, we consider a more suitable alternative, namely, the Frank-Wolfe algorithm \cite{reddi2016stochastic,miao2021learning,pokutta2020deep}, which is a projection-free first-order algorithm for constrained optimization.
Unlike Gradient Descent methods that rely on projection steps \cite{boyd2004convex}, the Frank-Wolfe algorithm employs a linear minimization oracle (LMO) \cite{reddi2016stochastic}: 
\begin{equation}
\boldsymbol{v}_t = LMO(\bigtriangledown \ell(\theta_t)) = \arg \min_{\boldsymbol{v} \in \mathcal{C}} <\bigtriangledown \ell(\boldsymbol{\theta}_t),\boldsymbol{v}>.
\end{equation}
Once $\boldsymbol{v}_t$ is obtained, the Stochastic Frank-Wolfe (SFW) update of $\boldsymbol{\theta}_t$ in the direction of $\boldsymbol{v}_t$ is given by:
\begin{equation}
\boldsymbol{\theta}_t\leftarrow \boldsymbol{\theta}_{t-1}+\eta\cdot(\boldsymbol{v}_t-\boldsymbol{\theta}_{t-1}).
\label{eq::theta_10}
\end{equation}
As in the case of SFw \cite{miao2021learning}, the LMO of $K$-polytope is given by: 
\begin{equation}
\boldsymbol{v}_i = \begin{cases}
-\lambda \cdot \text{sign}(m_i), & \text{if $\boldsymbol{m}_i$ is in the largest $K$ }\\
&\text{coordinates of $\boldsymbol{m}$}; \\
0, & \text{otherwise},
\end{cases}
\label{eq::v_11}
\end{equation}
where $\lambda$
represents the polytope radius and $\boldsymbol{m}_i$ can be regarded as a momentum vector \cite{boyd2004convex}, which leverages historical gradient information. 

\subsubsection{Overall optimization.}
The optimization of Equation 10 can be  decomposed into two parts.
First, we derive the update for the gradient $\boldsymbol{g}_{\boldsymbol{\theta}_t}$ of the parameter vector $\boldsymbol{\theta}$ using an approximation of SAM \cite{foretsharpness,kim2022fisher}:
\begin{equation}    
\boldsymbol{g}_{{\boldsymbol{\theta}_t}} = \sum_{i \in \mathcal{D}_t}\bigtriangledown_{\boldsymbol{\theta}_t}\ell_i(\boldsymbol{\theta}_t+\frac{\rho}{|\mathcal{D}_t|}\sum_{j \in \mathcal{D}_t}\bigtriangledown_{\boldsymbol{\theta}_t}\ell(\mathcal{D}_t;\boldsymbol{\theta}_t)).
\label{eq::updateing_10}
\end{equation}

Then we update the momentum vector based on the gradient $\boldsymbol{g}_{\boldsymbol{\theta}_t}$ and the learning rate $\alpha$:
\begin{equation}
    \boldsymbol{m}_t\leftarrow \boldsymbol{m}_{t-1}+\alpha \boldsymbol{g}_{\boldsymbol{\theta}_t}.
\end{equation}

We update the model parameters with Equation 12 through the calculation of LMO in Equation 13. 

\vspace*{-0.2cm}
\subsection{Intelligent Data Selection.}
We now proceed to introduce the second efficiency enhancement technique, DSCL (Data Selection for Continual Learning), which reduces the computational overhead of EsaCL by selecting a subset of informative training samples from $\mathcal{D}_t$. 
As discussed in Section 3.2, EsaCL aims to find a sharpness-aware sparse optimizer by maximizing inner perturbations and minimizing outer objective \cite{foretsharpness}.
Clearly, the training data $\mathcal{D}_t$ are utilized twice in both the inner and outer gradient steps. To address this, we propose two straightforward and efficient strategies to obtain an informative subset set of $\mathcal{D}_t$: support example selection (SSE) and redundant example elimination (ERE). These steps aim to reduce the amount of training data used without incurring significant reduction in performance of the resulting model.

\vspace*{-.1cm}
\subsection{Selection of support examples (SSE)}
As seen from Equation \ref{eq::max_ep_7}, we estimate the sharpness of the objective function with respect to the parameters through worst-case perturbation in the inner maximization step. Perturbation along the direction of $\boldsymbol{\epsilon}$ aims to maximize the average loss over the current batch in $\mathcal{B} \subset \mathcal{D}_t$. To further improve data efficiency, we aim to select a compact set of support examples $\mathcal{B}_s$ as well as to guarantee the upper bound in Equation \ref{eq::max_ep_7}, namely:
\begin{equation}
\mathcal{S}(\boldsymbol{\theta};\mathcal{B}) \approx  
\mathcal{S}(\boldsymbol{\theta};\mathcal{B}_s).
\label{eq::selection_upper_bound_11}
\end{equation}
Given the training loss over $\mathcal{D}_t$, we sort the training loss in descending order. Then, we select the support examples with the $Top\textit{-}J$ training loss values.
\begin{equation}
\mathcal{B}_s = \{(x_i,y_i) \in \mathcal{B}: i \in Top\textit{-}J( \ell(\theta;(x_i,y_i)),J \},
\label{eq::support_s_12}
\end{equation}
where $Top\textit{-}J(\boldsymbol{v}, J)$ returns the $Top\textit{-}J$ largest values in $\boldsymbol{v}$ and $J$ controls the size of $\mathcal{B}_s$. Formally, we can effectively obtain ascent gradients and achieve worst-case perturbation in parameter space by:
\begin{equation}
    \boldsymbol{\epsilon} = 
    \rho \cdot\bigtriangledown_{\theta}\ell(\mathcal{B}_s;\theta)/||\bigtriangledown_{\theta}\ell(\mathcal{B}_s;\theta)||_2.
\end{equation}
 
\subsection{Elimination of redundant examples (ERE)}
We have shown how to effectively choose support examples that significantly contribute to the worst-case perturbation. However, it is still computationally cumbersome to handle instance-wise gradients over $\mathcal{D}_t$ in the outer parameter minimization. Hence, we aim to eliminate to the extent possible the redundant samples in $\mathcal{D}_r$ without sacrificing the performance of the model.
Formally, given the perturbation $\boldsymbol{\epsilon}$,
our objective can be formulated as:
\begin{equation}
\mathbb{E}_{\mathcal{D}_t}\min_{\theta}\ell(\mathcal{\theta}+\boldsymbol{\epsilon} ) \simeq 
\mathbb{E}_{\mathcal{D}_t\setminus\mathcal{D}_r}\min_{\hat{\theta}}\ell(\mathcal{\hat{\theta}}+\boldsymbol{\epsilon} ),
\label{eq::data_selection_14}
\end{equation}
where $\mathcal{\theta}$ and $\mathcal{\hat{\theta}}$ (or $\hat{\mathcal{\theta}}_{-\mathcal{D}_r}$ ) denotes the parameters before and after redundant data pruning. 
Direct optimization of Equation \ref{eq::data_selection_14} is prohibitively expensive due to  Leave-One-Out estimation of impact of eliminating each sample. 
Hence, we propose to use estimated influence \cite{koh2017understanding} as an efficient approximation 
to understand model behavior over input space changes. 
When a training example is up-weighted by a small $\boldsymbol{\delta }$, the resulting impact on the parameters is given by:  
$\theta_{x,\delta } = \arg \min_{\theta}\frac{1}{n}\sum_{i=1}^{n}\\
\ell(\theta;x_i)+\delta\ell(\theta;x)$. The influence of weighting $x$ on parameters $\hat{\theta}$ is given by:
\begin{equation}
\mathcal{I}_{params}(x) = \frac{d \hat{\theta}_{x,\delta}}{d \delta}|_{\delta=0} =-H_{\hat{\theta}}^{-1}\bigtriangledown _{\theta}\ell(x,\theta),
\label{eq::delta_15}
\end{equation}
where $H_{\hat{\theta}} = \frac{1}{n}\bigtriangledown_{\theta}^2\ell(x,\theta)$ is the positive definite Hessian matrix.
Since removing an example is equivalent to assigning the value -$\frac{1}{n}$ to $\delta$, we can approximate parameter change without retraining on the data: $\hat{\theta}_{x}$ - $\hat{\theta}\approx -\frac{1}{n}\mathcal{I}_{params}(x)$.
However, the need to calculate the first-order gradient and Hessian matrix in Equation \ref{eq::delta_15} does not help reduce the time complexity of Equation \ref{eq::data_selection_14}.
Instead, we approximate the absolute value of parameter change as a result of removing $\mathcal{D}_r$ as: 
\begin{equation}
\begin{aligned}
||\hat{\theta}_{-\mathcal{D}_r}- \hat{\theta}||_2^2 
& \approx 
||\sum_{x_i \in \mathcal{D}_r}
-\frac{1}{n}
\mathcal{I}_{param}(x_i)||_2^2 \\
&=
||\sum_{x_i \in \mathcal{D}_r}
\frac{1}{n}
H_{\hat{\theta}}^{-1}\bigtriangledown _{\theta}\ell(x_i,\theta)||_2^2 \\
& \leqslant 
\sum_{x_i \in \mathcal{D}_r}
\frac{1}{n}
\frac{||\bigtriangledown_{\theta}\ell(x_i,\theta)||_2^2}{||H_{\hat{\theta}}||_2^2}.
\end{aligned}
\end{equation}
Once the Hessian matrix is calculated, the error in training loss resulting from elimination of the redundant data can be bounded by the magnitude of $\bigtriangledown _{\theta}\ell(x_i,\theta+\boldsymbol{\epsilon})$. 
This means samples with low first-order gradients have little influence on the loss changes. We simply approximate their gradients with the training loss itself, leveraging its local smoothness property. Hence, we can select the samples with the $Top\textit{-}J$ training loss values to contribute to the outer minimization step.
\section{Experiments}
We report results of comparison of EsaCL with 
baselines using several  continual learning benchmark data sets in the Task Incremental (Task-IL) and Class Incremental (Class-IL) learning settings.
\vspace{-0.2cm}
\subsection{Dataset Description.}
We used three continual learning benchmark datasets, namely, Split CIFAR-10, Split CIFAR-100 \cite{wangsparcl}and Tiny-ImageNet \cite{deng2009imagenet}.
Following \cite{wangsparcl,golkar2019continual},
We split Split CIFAR-10, Split CIFAR-100 and Tiny-ImageNet dataset into 5, 10 and 10 tasks, each task consists of 2, 10, 20 classes. 
\vspace*{-0.2cm}
\subsection{Baseline Algorithms}
We compare our EsaCL with several strong state-of-the-art CL baselines:
\begin{table*}
\centering
\caption{Performance in terms of effectiveness performance comparison on three benchmark datasets. ($\uparrow$) means the higher, the better. ($\downarrow$) means the lower, the better. The magnitude of training FLOPs is $10^{15}$.}
\resizebox{\textwidth}{!}{
\begin{tabular}{cc|ccc|ccc|ccc} 
\hline
\hline

\multirow{3}{*}{Method} 
& \multirow{3}{*}{Sparsity} 

&\multicolumn{3}{c|}{Split CIFAR-10} 
&\multicolumn{3}{c|}{Split CIFAR-100} 
& \multicolumn{3}{c}{Split Tiny-ImageNet} 
\\
& 
&\multirow{2}{*}{Task-IL ($\uparrow$)} 
& \multirow{2}{*}{CAP ($\downarrow$) } 
& Training 
& \multirow{2}{*}{Task-IL ($\uparrow$)} &\multirow{2}{*}{CAP ($\downarrow$)}  
& Training
& \multirow{2}{*}{Task-IL ($\uparrow$)} &\multirow{2}{*}{CAP ($\downarrow$)}  
& 
Training\\

& 
&  
&  
& FLOPs ($\downarrow$) 
&  
&   
& FLOPs ($\downarrow$)
&  
&   
& FLOPs ($\downarrow$)\\


\cline{1-11}
\cline{1-11}
La-MAML
& \multirow{5}{*}{-}
&$78.56_{\pm 0.74}$
&100
&1.8

& $71.37_{\pm 0.67}$ 
&100
&2.46

&$66.90_{\pm 0.10}$ 
& 100
&2.51
\\

GPM  
& 
& $79.41_{\pm 0.86}$ 
&100
&2.17
& $71.25_{\pm 0.32}$  
&100
&2.71
&$60.92_{\pm 0.10}$ 
& 100
&2.74
\\

LwF 
&
& $63.29_{\pm 0.12}$ 
& 100
& 1.7
& $46.37_{\pm 0.22}$ 
&100
&2.2
& 
$15.72_{\pm 0.69}$ 
& 100
& 2.5
\\

EWC  
&
& $67.24_{\pm 2.12}$ 
& 100 
& 1.7

&$58.80_{\pm 2.12}$   
&100
&2.2

&$19.58_{\pm 0.79}$ 
& 100
& 2.5\\

CO2L  
&
& $82.30_{\pm 2.13}$ 
& 100
&2.2

& $60.16_{\pm 0.32}$
&100
& 2.5

&$39.87_{\pm 0.54}$
& 100
&1.3 \\

\hline
SparCL-EWC
&\multirow{3}{*}{0.8}
&$68.33_{\pm 0.54}$
& 55.3
& 0.53

& $59.53_{\pm 0.25}$
& 59.7
&0.59

&$20.54_{\pm 0.62}$ 
& 65.4
& 3.2 \\

SparCL-CO2L
&
&$83.21_{\pm 0.57}$ 
& 60.1
& 0.6

&  $62.64_{\pm 0.33}$ 
& 67.3
&0.67

&$41.27_{\pm 0.52}$ 
& 64.1
& 0.8\\

\rowcolor{gray!40}EsaCL-EWC
&
&$69.02_{\pm 0.33}$ 
& 20
& 0.51

& $61.57_{\pm 0.61}$ 
& 20
& 0.53

&$22.54_{\pm 0.28}$ 
& 20
& 2.9 \\

\rowcolor{gray!40}EsaCL-CO2L
&
&$84.09_{\pm 0.21}$
& 20
& 0.6

&$62.91_{\pm 0.42}$ 
&20
&0.65

&$42.51_{\pm 0.53}$ 
& 20
& 0.6\\

\hline
PackNet 
& \multirow{3}{*}{0.50}
&$93.73_{\pm 0.55}$
& $86.7$
& 0.86

&$72.39_{\pm 0.37}$
&$96.3$
&1.42

& $60.46_{\pm 1.22}$
& $188.6$
& 1.94
\\

WSN 
& 
& $44.79_{\pm 1.86}$
&95.6
&0.91

& $76.38_{\pm 0.34}$
&$99.1$
&1.31

& $69.06_{\pm 2.35}$
& $92.1$
&1.56
\\

NISPA 
& 
& $57.36_{\pm 1.92}$ 
& 50
& 0.79

& $65.36_{\pm 2.19}$ 
&50
&1.29

& $59.56_{\pm 0.35}$ 
& 50
& 1.16 
\\

\hline
\hline
iCARL$^{500}$ 
&\multirow{3}{*}{0}
&$90.58_{\pm0.35}$ 
& 100
&1.92

&$63.22_{\pm 1.42}$ 
&100
&2.61

&$31.55_{\pm 3.27}$ 
& 100
&2.81
\\

ER$^{500}$ 
& 
&$ 93.61_{\pm0.27}$ 
& 100
& 1.92

&$68.23_{\pm0.17}$ 

&100
&2.37

&$48.64_{\pm0.46}$ 
& 100
&2.81
\\

DER++$^{500}$  
& 
&$93.88_{\pm 0.60}$ 
& 100 
&2.01

&$70.61_{\pm 0.08}$ 
&100
&2.43

&$51.78_{\pm 0.88}$ 
& 100
&2.96
\\

\hline
\hline
SparCL-ER$^{500}$
&\multirow{3}{*}{0.8}
&$92.96_{\pm 0.82}$ 
& 57.6
& 0.31

& $69.12_{\pm 0.22}$ 
& 45.6
& 0.68

&$50.35_{\pm 0.22}$ 
& 50.2
& 0.51\\

SparCL-DER++$^{500}$
&
&$94.12_{\pm 0.59}$ 
& 57.7
& 0.34

& $71.35_{\pm 0.41}$ 
& 62.1
& 0.62

&$51.97_{\pm 0.51}$ 
& 47.3
& 0.44
\\

\rowcolor{gray!50}EsaCL-ER$^{500}$
&
&$93.85_{\pm 0.56}$ 
& 20
& 0.32

& $70.51_{\pm 0.34}$ 
&20
&0.53

&$51.73_{\pm 0.55}$ 
& 20
& 0.46
\\
\rowcolor{gray!50}EsaCL-DER++$^{500}$
&
&$94.96_{\pm 0.68}$ 
& 20
& 0.33

& $72.94_{\pm 0.62}$ 
&20
&0.59

&$53.27_{\pm 0.92}$ 
& 20
& 0.43\\

\hline
MTL 
& 0.00 
& $97.96_{\pm 0.31}$ 
& 100
& 1.7

& $80.82_{\pm 0.30}$ 
&100
& 2.53

& 78.01
& 100
& 2.7
\\

\hline
\hline

\end{tabular}
}
\end{table*}
\begin{enumerate}
\vspace*{-0.1cm}
    \item[i]
{\bf Regularization-based methods}:
\circled{1} EWC \cite{de2021continual} constrains important parameters to stay close to their old values and leverages Fisher information matrix to measure the distance. 
\circled{2} LwF \cite{li2017learning} learns
new training samples while using their predictions from the
output head of the old tasks to compute the distillation loss.
\circled{3} GPM \cite{saha2021gradient} maintains the gradient subspace important to the old
tasks (i.e., the bases of core gradient space) for orthogonal
projection in updating parameters.
\vspace*{-0.1in}
\item[ii]{\bf Replay-based methods}:
\circled{1} ER\cite{chaudhry2019tiny} typically stores a
few old training samples within a small memory buffer.
\circled{2} DER++ \cite{buzzega2020dark} stores logits sampled throughout the
optimization trajectory, which resembles having several different teacher parameterization.
\circled{3} iCARL\cite{rebuffi2017icarl} stores a subset
of exemplars per class, which are selected to best approximate class
means in the learned feature space. At test time, the class
means are calculated for nearest-mean classification based
on all exemplars.
\vspace*{-0.1cm}
\item[iii]{\bf Architecture-based methods}: 
\circled{1} PackNet  \cite{kang2022forget} iteratively assigns parameter subsets to
consecutive tasks by constituting binary masks. 
\circled{2} NISPA \cite{gurbuz2022nispa} explicitly identifies the important neurons or
parameters for the current task and then releases the unimportant parts to the following tasks.
\circled{3} WSN\cite{kang2022forget} explicitly optimizes a binary mask to select dedicated neurons or
parameters for each task, with the masked regions of the
old tasks (almost) frozen to prevent catastrophic forgetting.
\vspace*{-0.1cm}
\item[iv] \textbf{Efficient continual learning methods}:
\circled{1} SparCL \cite{wangsparcl} is the most recent CL methods that addresses the efficiency considerations of continual learning.
Specifically, SparCL utilizes a pruning-retraining pipeline to iteratively enforces $l_1$ regularization to obtain a sparse model. 
\vspace*{-0.1cm}
\item[v] \textbf{Multi-task learning methods}: 
Multitask Learning (MTL) trains on multiple tasks simultaneously.
\end{enumerate}
\subsection{Evaluation Metrics}
We evaluate the performance of EsaCL from both effectiveness and efficiency dimensions.
To evaluate the classification performance, we use the $Acc_T$ metric, which is
the average of the final classification accuracy on all tasks \cite{de2021continual,gurbuz2022nispa}.
\vspace{-0.1cm}
\begin{equation}
 Acc_T = \frac{1}{T}\sum_{i=1}^{T}R_{T,i}.
\end{equation}

To evaluate its efficiency, we measure the training floating-point operations (FLOPs) \cite{wangsparcl}, model capacity \cite{kang2022forget} and memory footprint \cite{wangsparcl}.

\
\subsection{Results}
\vspace*{-0.5cm}
\subsubsection*{EsaCL versus Baselines.}
Table 2 presents the evaluation of different CL models on multiple continual learning benchmark datasets. The results show that:
\begin{enumerate}
\item[1]
Model sparsity helps enhance generalization ability in the continual learning setting. We see that SparCL-EWC and EsaCL-EWC outperform EWC with relative gains of $2\%$ and $3.5\%$ respectively on Split CIFAR-10 and Split CIFAR-100 datasets. 
\vspace*{-0.2cm}
\item[2] 
Across three datasets, EsaCL outperforms the regularization-based (SparCL-EWC and SparCL-CO2L) and the replay-based (SparCL-ER$^{500}$, SparCL-DER++$^{500}$)  variants of SparCL.
This finding validates our hypothesis (See Section 3) that flat loss landscape helps stabilize the training loss during pruning thus avoiding the need for post-pruning retraining of the model. 
\vspace*{-0.2cm}
\item[3] Both SparCL and EsaCL exhibit significant reductions in training FLOPs. However, SparCL requires additional space, almost double that used by EsaCL, because of its need to store intermediate outputs, such as generated masks, in each iteration. 
\vspace*{-0.2cm}
\item[4] As a typical strategy of architecture-based methods, NISPA can also enhance memory efficiency due to its low capacity requirements. 
However, NSPA is highly sensitive to parameter tuning and exhibits limited performance. In contrast,  EsaCL offers a simple, yet highly effective and easy to implement approach to efficient continual learning.
\end{enumerate}
\vspace*{-0.3cm}
\subsection{Detailed Analysis}
We proceed to present a comprehensive analysis of EsaCL and its variants. 
Unless otherwise specified, we use "EsaCL" to denote the EsaCL-DER++$^{500}$ method.
EsaCL has two components: Sharpness-aware Pruning and Intelligent Data Selection. 
Next, we discuss the results of our sensitivity analysis of  
sparsity-ratio and user-specified parameters. 

\begin{table}[ht]
\begin{center}
 \caption{Ablation study of sharpness-aware informed pruning method on CIFAR-100 dataset.}
 \resizebox{\linewidth}{!}{
\begin{tabular}{ c c c c } 

\hline
\hline
\multirow{2}{*}{Sparsity Ratio} 
&\multirow{2}{*}{Variants}  &\multirow{2}{*}{Class-IL ($\uparrow$)} 
&Memory\\
&&& footprints ($\downarrow$)\\
\hline
 \multirow{3}{*}{0.6} &Ours w/o SA & 34.72 &  172MB\\  
 &Ours w/o SP & 47.01 & 227MB \\
 &Ours & 50.82 & 181MB \\
 \hline
  \multirow{3}{*}{0.7} &Ours w/o SA & 31.29 & 172MB \\  
 &Ours w/o SP & 45.35 & 227MB \\
 &Ours & 50.49 & 181MB \\
 \hline
  \multirow{3}{*}{0.8} &Ours w/o SA & 26.51 &  172MB\\  
 &Ours w/o SP & 41.12 & 227MB \\
 &Ours & 49.56 & 181MB \\
 \hline
  \multirow{3}{*}{0.9} &Ours w/o SA & 20.86 & 172MB \\  
 &Ours w/o SP & 34.81 & 227MB \\
 &Ours & 45.37 & 181MB \\
 \hline
 \hline
\label{tab::ablation_pruning}
\end{tabular}
}
\vspace{-1cm}
\end{center}
\end{table}

\subsubsection*{Ablation study of sharpness-aware pruning.}
We conducted an ablation study to assess the impact of the sharpness-aware sparse optimizer. We compared our method with two variants: "EsaCL w/o SA" (without sharpness-aware constraint in Equation 9) and "EsaCL w/o SP" (without sparse constraint in Equation 9). The results of this ablation study, including various sparsity ratios, are presented in Table \ref{tab::ablation_pruning} on the CIFAR-100 dataset. The 
Table \ref{tab::ablation_pruning} presents the results of these experiments on the CIFAR-100 dataset, for several different choices of the sparsity ratio.

First, we observe that the dropping either SA or SP results in a substantial degradation in performance. This suggests that both sharpness-aware minimization and sparse projection contribute to improvements in efficiency and accuracy, across a broad range of sparsity ratios. Additionally, comparing our method without SA (EsaCL w/o SA) to the EsaCL, we find a slightly lower memory footprint when a sharpness-aware minimizer is not used. This could be because the sparse projector encourages model sparsity even without SA. However, it is evident that the inclusion of the sharpness-aware minimizer leads to a significant performance increase, which validates our initial hypothesis.
We further note that the performance decline in EsaCL w/o SA is more pronounced compared to EsaCL w/o SP. This underscores the importance of finding a flat loss landscape to ensure optimal model performance. Moreover, the benefits of SA and SP appear to be additive and hence mutually complementary.


\begin{figure}[th]
  \centering
  \begin{subfigure}[b]{0.48\linewidth}
\includegraphics[width=0.95\textwidth]{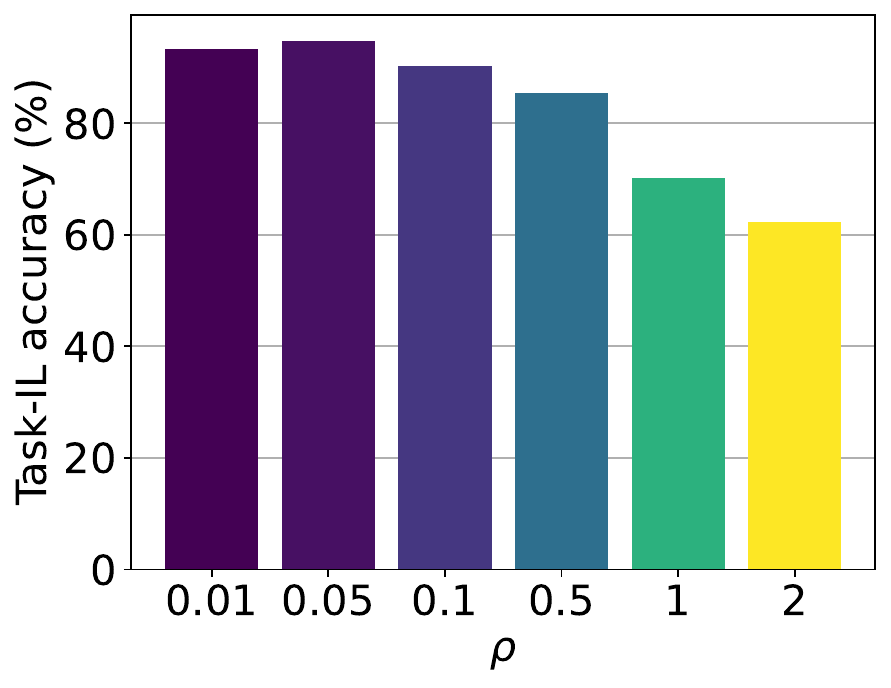}
\caption{CIFAR-10 dataset.}
\end{subfigure}
\begin{subfigure}[b]{0.48\linewidth}
\includegraphics[width=0.9\textwidth]{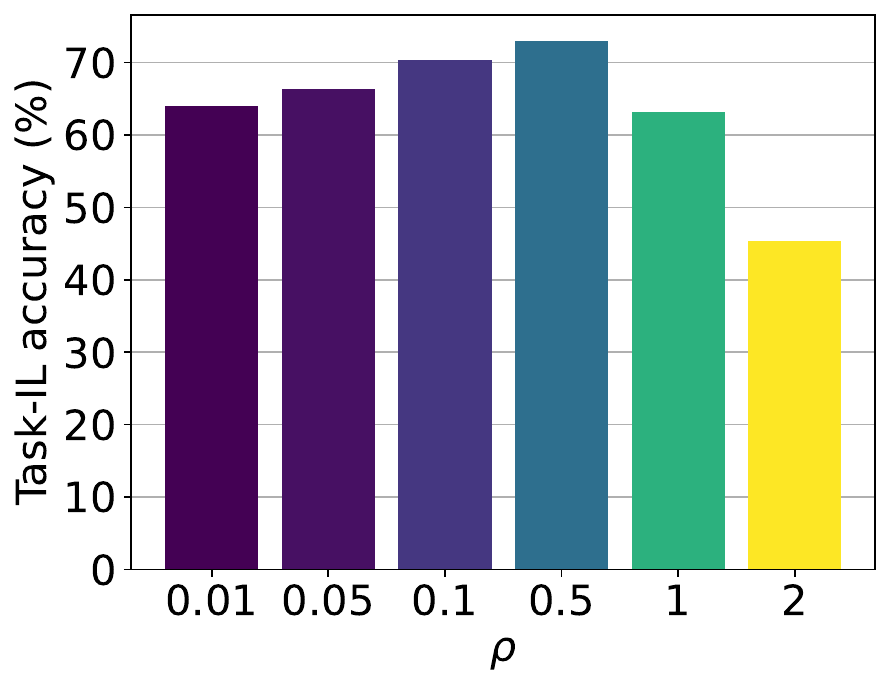}
    \caption{CIFAR-100 dataset.}
\end{subfigure}
\caption{Task incremental average accuracy w.r.t. weight perturbation ratio $\rho$.}
\label{Fig:rho_3}
\end{figure}
\vspace{-0.3cm}
\begin{figure}[h]
  \centering

  \begin{subfigure}[b]{0.48\linewidth}
\includegraphics[width=0.95\textwidth]{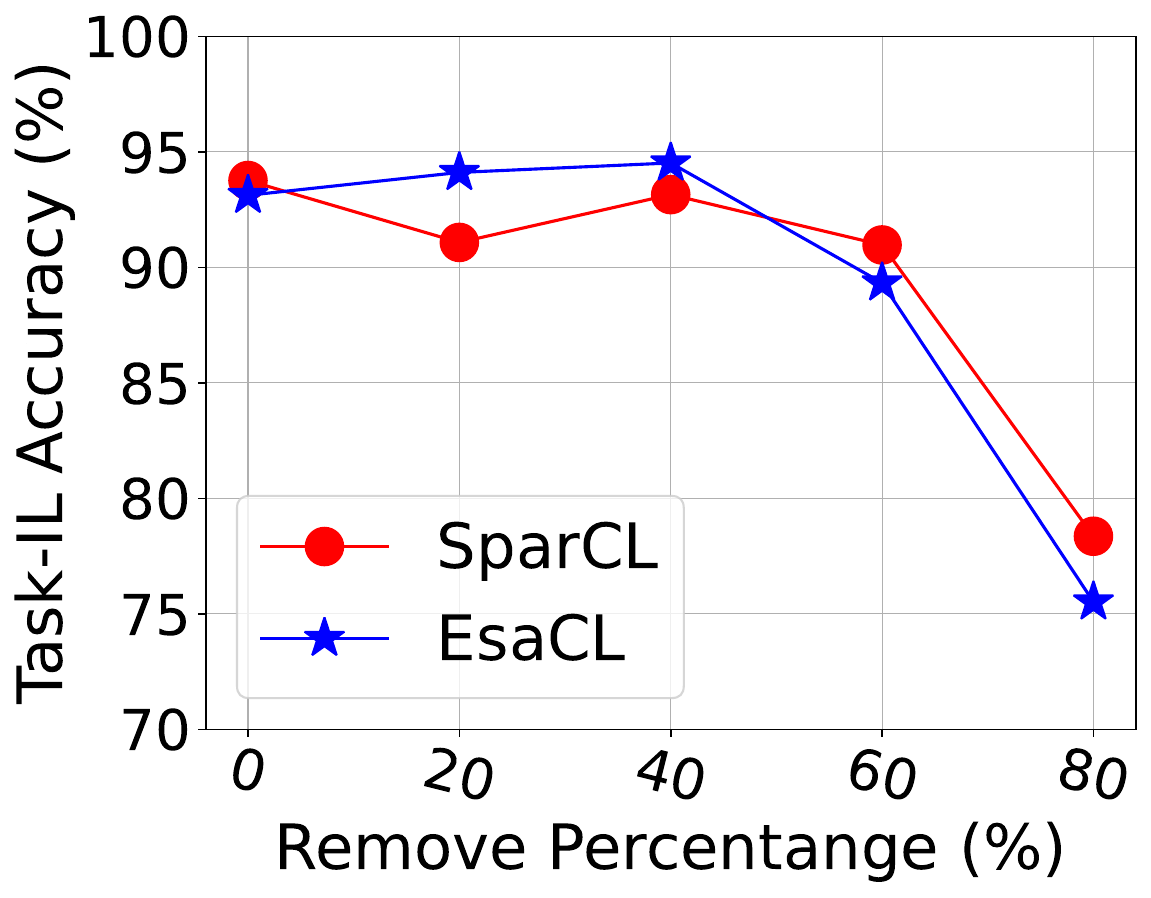}
\caption{CIFAR-10 dataset.}
\end{subfigure}
\begin{subfigure}[b]{0.48\linewidth}
    \includegraphics[width=0.9\textwidth]{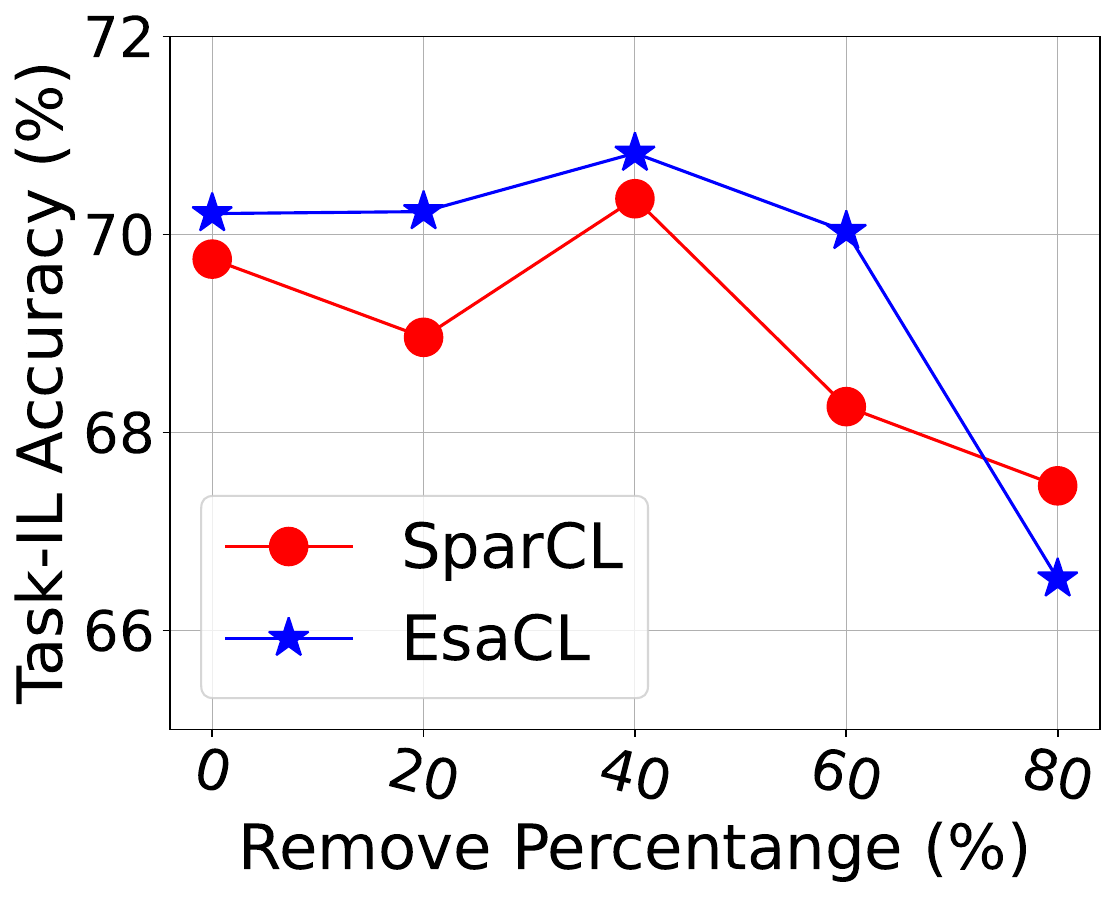}
    \caption{CIFAR-100 dataset.}
\end{subfigure}
\caption{Comparison between SparCL and EsaCL data removal strategy w.r.t. different data removal proportions.}
\vspace{-0.3cm}
\label{Fig:parameter_sensitivity}
\end{figure}
\vspace{-0.3cm}
\subsubsection*{Ablation study of intelligent data selection.}
\begin{table}
\begin{center}
\caption{Ablation study of Intelligent Data Selection method on three benchmark datasets.}

\resizebox{\linewidth}{!}{
\begin{tabular}{cccc} 
\hline
\hline 

\multirow{2}{*}{Dataset}
& \multirow{2}{*}{Variants} 
&\multirow{2}{*}{Task-IL ($\uparrow$)} 
& Training \\
&
&
& FLOPs ($\downarrow$) \\
\hline
 \multirow{3}{*}{Split CIFAR-10}& Ours w/o SSE
& 93.92
& 0.46\\

& Ours w/o ERE
& 95.37
&0.39 \\

& Ours
& 94.81
& 0.33\\
\hline
 \multirow{3}{*}{Split CIFAR-100}& Ours w/o SSE
&  72.01
& 0.82\\

& Ours w/o ERE
&73.62
& 0.67\\

& Ours
& 72.53
& 0.59 \\
\hline
 \multirow{3}{*}{Tiny-ImageNet}& Ours w/o SSE
&  52.23
& 0.71\\

& Ours w/o ERE
&53.09
&0.48\\

& Ours
& 53.16
& 0.43\\
\hline
 \hline
\label{tab::ablation_data}
\end{tabular}
}
\vspace{-1.1cm}
\end{center}
\end{table}

We conducted ablation experiments to analyze the contribution of data selection techniques in the context of task incremental learning. 
Table \ref{tab::ablation_data} clearly shows that both data selection methods  significantly reduce training FLOPs across the three benchmark datasets, with concomitant gains improving training efficiency. Notably, the SSE techniques reduce FLOPs from 0.46 to 0.33 on the CIFAR-10 dataset
When combining ERE with the data selection methods, we observe an improvement in classification performance on both the CIFAR-10 and CIFAR-100 datasets. However, it is important to note that this improvement comes at the expense of a slight increase in training FLOPs.
An interesting observation is that on the Tiny-ImageNet dataset, removing the ERE component results in a slight decrease in performance. 
This could be because noise in this data set is large enough to negatively affects the model performance, and IDS component indirectly improves training data quality, and hence the performance as well.

\vspace*{-0.2cm}
\subsubsection*{Parameter Sensitivity Analysis.}
We conducted experiments to examine how the choices of $\rho$ and $\mathcal{B}_s$ influence the performance of EsaCL.
Here, $\rho$ denotes the radius of perturbation, and $\mathcal{B}_s$ denotes the number of selected data.
The results of these experments are
shown in Figure \ref{Fig:rho_3} and Figure \ref{Fig:parameter_sensitivity}, respectively. 
From Figure \ref{Fig:rho_3} (b), it is evident that the classification performance gradually increases as $\rho$ ranges from 0.01 to 0.5 on the CIFAR-100 dataset. This observation aligns with our motivation that a flat loss landscape can enhance the model's generalization ability. However, a sudden increase in the value of $\rho$ leads to a drop in performance from $72.93\%$ to $45.31\%$. As mentioned in Section 3.1, the flat region is confined to a relatively small neighborhood around the current optimum, indicating that the optimal value of $\rho$ should be bounded within a certain range.
Figure \ref{Fig:parameter_sensitivity} (b) clearly shows that SparCL suffers a drop in performance when $20\%$ of the data are removed, but achieves its peak performance when   $40\%$ of the data is removed. In comparison, EsaCL exhibits greater robustness to varying levels of data removal, particularly when the parameter B increases from 0 to $40\%$.

\section{Summary and Discussion}
In this paper, we have introduced EsaCL, that incorporates two novel approaches to increasing the efficiency of training in the continual learning setting: sharpness-informed directional pruning (SDP) and intelligent data selection (IDS). 
Experiments on standard CL benchmarks show that EsaCL achieves performance that is competitive with the SOTA methods, while requiring less memory and computational resources.
Some promising directions for future work include improved methods for data subset selection, improved methods for searching for and optimizing over the flat region of the objective function, and rigorous theoretical analyses of EsaCL.
\vspace{-0.3cm}
\subsubsection*{Acknowledgments.}
This work was funded in part by grants from the National Science Foundation (2226025, 2041759), and the National Center for Translational Sciences of the National Institutes of Health (UL1 TR002014).

\newpage
\bibliographystyle{siam}

\bibliography{reference}

\newpage
\begin{table*}[h]
\centering
\caption{Classification accuracy in terms of different sparisity ratio on CIFAR 100 dataset. ($\uparrow$) means the higher, the better.
Notably, SparCL-ER$^{500}$ follows a `pruning-retraining' pipeline and our EsaCL-ER$^{500}$ is a one-shot pruning method that eliminates the need for the retraining stage.}
\begin{tabular}{cc|cccc} 
\hline
\hline

\multirow{2}{*}{Method} 
& \multirow{2}{*}{Sparsity} 
&\multicolumn{4}{c}{Split CIFAR-100}  
\\
& &\multirow{2}{*}{Task-IL ($\uparrow$)} 
& \multirow{2}{*}{Class-IL ($\uparrow$)}  
& Training
& Capacity \\
&&&&&Time\\
\\ 
\cline{1-6}

\hline
SparCL-ER$^{500}$ &\multirow{2}{*}{0.60} 
&$69.75_{\pm 0.68}$ 
& $49.88_{\pm 0.72}$&
23h
& 83.3
\\
EsaCL-ER$^{500}$ 
&
&$70.21_{\pm 0.68}$ 
& $50.79_{\pm 0.72}$
& 16h
& 40.0
\\
\hline

SparCL-ER$^{500}$ &\multirow{2}{*}{0.70} 
&$68.96_{\pm 0.68}$ & 
$48.12_{\pm 0.72}$
& 23h
& 77.2
\\
EsaCL-ER$^{500}$ &
&$70.36_{\pm 0.68}$ 
& $50.05_{\pm 0.72}$
& 16h
&30.0
\\
\hline

SparCL-ER$^{500}$ &\multirow{2}{*}{0.80} 
&$70.59_{\pm 0.68}$ 
& $48.35_{\pm 0.72}$
& 23h
& 45.6
\\

EsaCL-ER$^{500}$ &
&$ 70.23_{\pm 0.68}$ 
& $49.27_{\pm 0.72}$
& 16h
& 20.0
\\
\hline

SparCL-ER$^{500}$ &\multirow{2}{*}{0.85} 
&$70.03_{\pm 0.68}$ & 
$47.32_{\pm 0.72}$& 23h
&15.0
\\
EsaCL-ER
$^{500}$ 
&&$69.51_{\pm 0.68}$ 
& $47.92_{\pm 0.72}$
& 16h
&15.0
\\
\hline

SparCL-ER$^{500}$ 
&\multirow{2}{*}{0.90} 
&$67.36_{\pm 0.68}$ 
& 
$46.51_{\pm 0.72}$
& 23h
& 25.2
\\
EsaCL-ER$^{500}$ 
&
&$66.22_{\pm 0.68}$ 
& $45.13_{\pm 0.72}$
& 16h
&10.0

\\
\hline
\hline
\label{tab::sparsity_ratio}
\end{tabular}
\end{table*}
\section{Appendix}
\subsection{Additional Details}
All baselines use stochastic gradient descent (SGD) as the underlying optimization method for training the respective models. Because we directly implement our method from the official code in SparCL \cite{wangsparcl}, we report the performance of SparCL using its default settings.
We set the per task training epochs to 50, 80 and 100 for Split CIFAR-10, Split CIFAR-100 and TinyImageNet dataset, respectively.
For the replay methods, we use memory buffer 500 for the three benchmarks as suggested in \cite{rebuffi2017icarl,wangsparcl}.
Mask matrix compression rate $\gamma$ in Eq.\ref{capacity} is set as 0.75 for all corresponding baselines.
$\tau$ controls the magnitude, which together influence the distribution of
the learned parameters.
We set $\tau$ as 5, 8, 8 and K as 0.05, 0.1, 0.08 in Split CIFAR-10, Split CIFAR-100 and TinyImageNet dataset, 
 respectively.
Learning rate is set as 0.01, 0.1, 0.01, respectively.
Besides, $\rho$ denotes the perturbation radio in SAM, which is set as 0.05, 0.5 and 0.1. For these three datasets, We select top $40\%$ samples through the method proposed in section 3.4.
All our experiments run on a single-GPU setup of NVIDIA V100.

\subsection{Experiments}
\subsubsection{Sparsity-ratio sensitivity Analysis}
Here we present experimental results to demonstrate the robustness of our method across different sparsity ratios. Due to space constraints, we focus on showcasing the results on the CIFAR-100 dataset. The experimental findings are summarized in Table \ref{tab::sparsity_ratio}.
First, we observe that our EsaCL framework achieves comparable accuracy to the SparCL method, even without the need for extra retraining, across a range of sparsity ratios.
Specifically, EsaCL holds a stable performance when sparsity ratio increases from $60\%$ to $80\%$.
Additionally, we observe that the  EsaCL  model is compact compared to its SparCL counterpart. This difference in model size arises from the inclusion of a learned mask indicator for each task in the SparCL framework, which introduces an efficiency burden.
\subsection{Evaluation Metrics}
In order to calculate FLOPs needed for a single forward pass of a sparse model, we count the total number of multiplications and additions layer by layer for a given layer
sparsity. The total FLOPs is then obtained by summing
up all of these multiply-add operations. We can calculate the training FLOPs as:
\begin{equation}
\small
\begin{aligned}
  &FLOPs =  \text{ (add-multiplies per forward pass)}\\
  &* \text{(2 FLOPs/add-multiply)}* \text{(examples numbers)}\\
  &* \text{(3 for forward and backward pass)} * \text{(epochs numbers)}.\\
  \end{aligned}
\end{equation}
Model capacity \cite{kang2022forget} measures the total of percentage of nonzero weights plus the compressed masks matrix for all tasks as follows:
\begin{equation}
\text{Capacity} = (1 - S) + \frac{(1-\gamma)T}{32},
\label{capacity}
\end{equation}
where S is the sparsity ratio and $\gamma$ is the mask matrix compression rate through 7bit Huffman encoding.

Following existing works \cite{wangsparcl,kang2022forget}, we consider memory footprints as consisting of three components: 1) model parameter allocations during training stage; 2) model intermediate outputs, i.e., activation; 3) parameter gradients. The memory footprint results are calculated by an approximate summation of these components.

\end{document}